\documentclass{article}

\PassOptionsToPackage{numbers, compress}{natbib}
\usepackage[preprint]{neurips_2026}
\usepackage[utf8]{inputenc} 
\usepackage[T1]{fontenc}    
\usepackage{hyperref}       
\usepackage{url}            
\usepackage{booktabs}       
\usepackage{amsfonts}       
\usepackage{nicefrac}       
\usepackage{microtype}      
\usepackage{xcolor}         
\usepackage{graphicx}
\usepackage{amsmath}
\usepackage{amssymb}
\usepackage{mathtools}
\usepackage{amsthm}
\usepackage{algorithm}
\usepackage{algorithmic}
\usepackage{wrapfig}

\title{PhysTacGen: Physics-Aware Visual-Tactile Sensor Image Generation}

\author{%
  Guo Tang\textsuperscript{1} \quad Yongtao Wang\textsuperscript{1,}\textsuperscript{2}\thanks{Corresponding author.}  \\
  \textsuperscript{1}Wangxuan Institute of Computer Technology, Peking University \\
  \textsuperscript{2} VGI Labs Co., Ltd. \\
  \texttt{confection\_tang@stu.pku.edu.cn} \\
  \texttt{wyt@pku.edu.cn}
}

\begin{document}

\maketitle

% ==============================================================================
% Abstract
% ==============================================================================

\begin{abstract}
Realistic physical interaction is a cornerstone of embodied intelligence, yet collecting paired visual--tactile data remains costly. Visual-to-tactile synthesis offers a promising approach to augmenting such data, but learning this mapping is complicated by the gap between visual appearance and contact-related material properties, as well as spatial misalignment in paired observations. To address these challenges, we present \textbf{PhysTacGen}, a visual-to-optical-tactile image generation framework that integrates material-aware descriptions with geometric conditioning. First, we introduce Group Tactile Policy Optimization (GTPO), a reinforcement learning strategy that refines a vision--language model to generate structured material descriptions using task-specific rewards. Second, we combine DINOv2-based pair curation with monocular relative-depth estimation to select training pairs and provide geometric priors. Finally, an SDXL ControlNet synthesizes optical tactile images conditioned on RGB, relative depth, and GTPO-generated text. Experiments on curated SSVTP data demonstrate improved structural similarity over the compared baselines, while a blinded user study shows a preference for GTPO-generated descriptions. Generated tactile inputs also improve performance on an attribute-derived force-coefficient prediction proxy. Together, these results demonstrate the effectiveness of PhysTacGen for optical tactile image synthesis and its utility in the evaluated downstream task.The code will be available at https://github.com/VDIGPKU/PhysTacGen.
\end{abstract}

% ==============================================================================
% 1. Introduction
% ==============================================================================
\section{Introduction}
\label{sec:intro}

High-precision manipulation is the cornerstone of next-generation embodied intelligence. Tasks requiring sub-millimetre accuracy---such as threading a cable, inserting a USB connector~\citep{zhang2026vtla}, or handling deformable tissues~\citep{heng2025vitacformer}---often render vision insufficient due to occlusion or lack of contact information. Consequently, tactile sensing has emerged as an imperative modality for the last centimetre of interaction, providing the temporal feedback and subtle surface deformation cues necessary for complex reasoning~\citep{feng2025anytouch}. The recent surge in Vision-Tactile-Language Models (VTLMs) further underscores the necessity of integrating touch with semantic understanding to achieve robust robot perception~\citep{zhang2026vtla}. However, a critical bottleneck impedes this progress: the scarcity of large-scale, paired visual-tactile data. Unlike the web-scale abundance of visual data or the maturity of photorealistic visual rendering, collecting high-fidelity tactile datasets is prohibitively expensive, hardware-dependent, and labour-intensive. 
% This data vacuum highlights an urgent need for scalable methods to synthesise realistic tactile responses from widely available visual observations.
% Beyond the absolute scarcity of samples, a more subtle but pervasive issue is the spatial misalignment caused by the varying installation positions of external cameras and tactile sensors. As illustrated in Figure 1, this lack of co-axial alignment makes it difficult for models to learn direct pixel-to-pixel correspondences, further complicating the synthesis of consistent tactile responses from visual observations.
Furthermore, as highlighted in recent multi-embodied studies \citep{li2025v}, the spatial relationship between external cameras and tactile sensors varies significantly across different robotic platforms. This inconsistency in camera-to-sensor layout introduces severe spatial noise into existing datasets, making it difficult for conventional models to learn a robust mapping from 2D visual pixels to 3D tactile deformations.
% \begin{wrapfigure}{r}{0.5\textwidth}
%   \begin{center}
%     \includegraphics[width=0.48\textwidth]{figure.pdf}
%   \end{center}
%   \caption{xxxxxxxxxxxxxxxxxx}
%   \label{fig:half_width_fig}
% \end{wrapfigure}
To address this scarcity, prior works have primarily turned to physical simulation. Simulators such as Tacto~\citep{wang2022tacto}, Taxim~\citep{si2022taxim}, and TacSL~\citep{akinola2025tacsl} leverage rendering engines to generate synthetic tactile images. While scalable, these approaches suffer from a fundamental domain gap. The rendered visual counterparts in simulators often lack the photorealism and diverse textures of real-world objects. Furthermore, the simplified contact physics in simulators frequently fails to capture the complex, non-linear deformations of elastomers found in real optical tactile sensors.

Alternatively, data-driven Visual-to-Tactile (V2T) synthesis has gained traction. Approaches like SingleS2R~\citep{tang2024singles2r} attempt to map visual features directly to tactile responses using pixel-level statistics. However, these methods expose two critical failure modes. First, they fail to distinguish between \textit{visual texture} and \textit{physical compliance}. A model might erroneously generate deep tactile deformations for a hard, marble-patterned surface simply because the visual texture is complex, or conversely, generate a flat response for a texture-less but soft sponge. Second, real-world visual-tactile pairs suffer from severe spatial misalignment due to mechanical tolerances. For deterministic pixel-wise MSE regression, uncertainty over spatial correspondence can average out detail; stochastic generators may instead produce sharp but misplaced features. Without explicit physical grounding and spatial alignment, these generative models hallucinate interactions that violate physical laws.

In this paper, we introduce \textbf{PhysTacGen}, a physics-aware generation framework that bridges the gap between visual appearance and tactile mechanics. PhysTacGen moves beyond simple modality mapping by explicitly modeling physical reasoning, filtering spatial noise, and injecting geometric constraints. On the evaluated curated SSVTP split, our method achieves the highest SSIM among the compared systems; the downstream experiment tests substitution of generated tactile images on a heuristic proxy task rather than robot transfer.

These three components form a cohesive pipeline: \textbf{GTPO} acts as the 'physical brain' to infer latent properties from semantic cues; the \textbf{Data Curation \& Prior} stage functions as the 'geometric filter' to resolve spatial noise; and finally, the \textbf{SDXL-based Synthesis} serves as the 'integrator' that fuses these semantic and geometric insights into high-fidelity tactile images.
To sum up, our system makes three core technical contributions:

\begin{itemize}
    \item \textbf{Group Tactile Policy Optimization (GTPO) for Physical Grounding.} We propose GTPO, a novel alignment approach inspired by reasoning models. GTPO utilizes Reinforcement Learning with a specialized suite of reward functions to refine Large Multimodal Models. By rewarding reasoning chains that align with contact mechanics and tribology, we generate precise, physics-informed textual descriptions.
    
    \item \textbf{Semantic-Driven Data Curation and Spatial Visual Prior.} To resolve the blurriness caused by spatial misalignment, we introduce a rigorous data curation pipeline using Foundation Models (DINOv2) to filter out geometrically mismatched pairs. Concurrently, we utilize monocular depth estimation to provide a relative macro-geometric prior alongside visual texture.
    
    \item \textbf{Physics-Driven Multi-Modal Synthesis via SDXL.} We propose a generative architecture powered by Stable Diffusion XL (SDXL) that fuses semantic reasoning with geometric guidance. We utilize the physics-rich text generated by GTPO as the prompt for the dual text encoders, while employing the RGB image and its estimated depth map as 4-channel structural conditions for high-frequency detail generation.
\end{itemize}

% ==============================================================================
% 2. Related Work
% ==============================================================================
\section{Related Work}
\label{sec:related}

\paragraph{Optical Tactile Sensing and Simulation.}
High-resolution optical tactile sensors, such as GelSight~\citep{yuan2017gelsight} and Digit~\citep{lambeta2020digit}, have revolutionized robotic manipulation by providing dense geometric information.
Recent advancements have further integrated multimodal fingertips to digitize touch~\citep{lambeta2024digitizing}. 
To mitigate the high cost of real-world data collection, simulators like Tacto~\citep{wang2022tacto}, Taxim~\citep{si2022taxim}  and TacSL~\citep{akinola2025tacsl} leverage rendering engines to synthesize tactile images. 
Other efforts utilize path tracing and IMPM to simulate complex dynamic contact like slip and rotation~\citep{shen2024simulation}. However, these analytical simulators often rely on simplified contact models that fail to capture the complex, non-linear deformations of elastomers found in real-world environments.

\paragraph{Tactile Representation and Foundation Models.}
With the advent of self-supervised learning, models like Sparsh~\citep{higuera2024sparsh} and AnyTouch ~\citep{feng2025anytouch} have established unified representations across diverse sensors. These frameworks often leverage vision foundation models such as MAE~\citep{he2022masked} or multi-modal alignment strategies like CLIP~\citep{radford2021learning} to bind touch to everything~\citep{yang2024binding}. Specifically, CLTP ~\citep{ma2026cltp} focuses on 3D contact geometry understanding, while Octopi~\citep{yu2024octopi} utilizes Large Multimodal Models for object property reasoning. Our work builds on this trend but shifts the focus from representation to generation, specifically addressing the semantic-physical gap that representation-only models might overlook.

\paragraph{Visual-to-Tactile (V2T) Synthesis and Force Perception.}

Mapping external RGB to optical tactile observations has been studied using image translation and reference-conditioned synthesis. Pix2Pix~\citep{isola2017image} provides a general paired translation baseline; Pix2Pix-Turbo~\citep{parmar2024one} uses a one-step text-to-image backbone. VisGel~\citep{li2019connecting} uses a query RGB image and a reference RGB--tactile pair, whereas Controllable Visual-Tactile Synthesis~\citep{gao2023controllable} generates visual and tactile outputs from editable sketches. The latter uses a different input and output protocol from our local optical-tactile prediction. Latent Diffusion Models~\citep{rombach2022high}, PolyTouch~\citep{zhao2025polytouch}, SingleS2R~\citep{tang2024singles2r}, and contact-force estimation~\citep{shahidzadeh2025feelanyforce} provide further context, but the experiments below assess tactile image synthesis and a heuristic coefficient proxy, not measured-force control.

\paragraph{Tactile-Aware Manipulation and Real-world Challenges.}
Dynamic tactile perception is essential for contact-rich tasks, as explored in Reactive Diffusion Policy~\citep{xue2025reactive} and ViTacformer~\citep{heng2025vitacformer}. The integration of language has led to the emergence of Vision-Tactile-Language-Action (VTLA) models~\citep{zhang2026vtla}. Despite these advances, real-world data collection remains plagued by spatial misalignment between cameras and sensors, especially in hardware-independent interfaces like Fast-UMI~\citep{wu2024fast}. Large-scale datasets such as SSVTP~\citep{kerr2022self} highlight the need for robust cross-modal alignment. Our framework contributes to this ecosystem by generating spatially aligned, physics-aware tactile data, motivating future validation in real manipulation tasks.

\begin{figure*}[t]
  \centering
  \includegraphics[width=\textwidth]{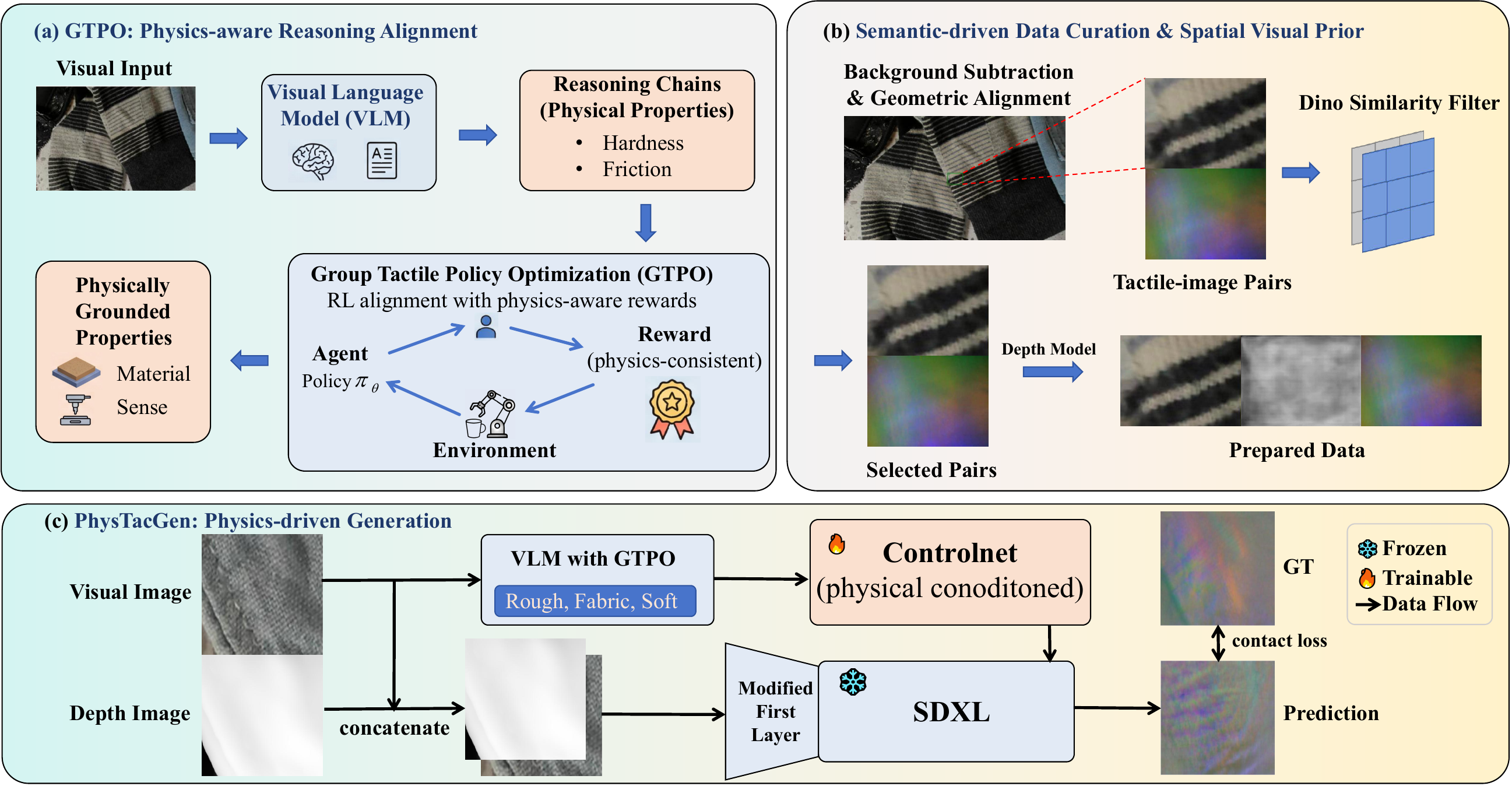}
  \caption{\textbf{The PhysTacGen Framework.} Given a visual observation, our system operates in three sequential stages: (1) \textbf{Physics-Aware Reasoning:} A VLM fine-tuned via \textit{GTPO} infers latent physical properties. (2) \textbf{Data Curation \& Spatial Prior:} The dataset is semantically purified using DINOv2, and input RGB is depth-estimated to form a 4-channel geometric condition. (3) \textbf{Multi-Modal Synthesis:} A physics-driven SDXL ControlNet synthesises the high-fidelity tactile sensor image $I_t$. Here $I_v$ is the external visual RGB image, $I_d$ the estimated relative-depth prior, and $c_{\mathrm{phys}}$ the GTPO-generated physical description; the optical tactile image is $I_t$.}
  \label{fig:pipeline}
\end{figure*}

% ==============================================================================
% 3. Method
% ==============================================================================
\section{Method}
\label{sec:method}
The goal of PhysTacGen is to synthesise high-fidelity tactile sensor images $I_t \in \mathbb{R}^{H \times W \times 3}$ given a visual observation $I_v \in \mathbb{R}^{H \times W \times 3}$. As illustrated in Figure~\ref{fig:pipeline}, the framework operates in three sequential stages to explicitly disentangle \textit{semantic physical reasoning} from \textit{geometric deformation generation}. Optical tactile RGB records a deformed, internally illuminated sensor membrane, rather than the contacting object\'s visible color; its color-to-gradient mapping depends on the sensor calibration. The RGB input $I_v$ yields both the physical description $c_{\mathrm{phys}}$ and estimated relative depth $I_d$, which condition the generator.

\subsection{Group Tactile Policy Optimisation (GTPO)}
\label{sec:gtpo}
Standard Supervised Fine-Tuning (SFT) often leads to superficial alignment where models mimic the style of training captions without grounding the underlying physics. Traditional RLHF requires training a critic model, which is extremely expensive and unstable in multimodal large models since the value model also needs to look at images, leading to memory explosion. To address this, we propose \textbf{Group Tactile Policy Optimisation (GTPO)}, a reinforcement learning framework built upon GRPO~\citep{shao2024deepseekmath}. However, standard GRPO is primarily designed for general semantic alignment or mathematical reasoning and lacks the domain-specific physical common sense required for tactile tasks. It cannot inherently parse the subtle relationship between visual textures and latent material mechanics. To bridge this gap, we specialize the framework into GTPO, which adapts the group-based reward mechanism to the nuances of tactile physics. GTPO refines the Large Multimodal Model (LMM) to infer latent physical properties from visual cues by rewarding reasoning chains that strictly adhere to physical laws. Its primary objective is to serve as a semantic physical interpreter, which guides the subsequent image synthesis process.

\paragraph{Objective Function.}
Given a visual prompt $x$, we sample a group of $G$ outputs $\{y_1, y_2, \dots, y_G\}$ from the current policy $\pi_\theta$. Unlike PPO, GRPO eliminates the need for a critic model by using group-based advantage estimation. The optimization objective is formulated as:
\begin{subequations}
\label{eq:gtpo_decomposed}
\begin{align}
    \rho_i(\theta) &= \frac{\pi_\theta(y_i|x)}{\pi_{\theta_{\text{old}}}(y_i|x)},  \\
    \mathcal{L}_i(\theta) &= \min \Big( \rho_i \hat{A}_i, \text{clip}(\rho_i, 1-\epsilon, 1+\epsilon) \hat{A}_i \Big),  \\
    \mathcal{J}(\theta) &= \mathbb{E}_{\mathcal{B}} \Bigg[ \frac{1}{G} \sum_{i=1}^G \Big( \mathcal{L}_i(\theta) - \beta \mathbb{D}_{\mathrm{KL}}(\pi_\theta || \pi_{\mathrm{ref}}) \Big) \Bigg], \label{eq:final_objective}
\end{align}
\end{subequations}
where $\rho_i$ is the importance sampling ratio, $\epsilon$ is the clipping coefficient, and $\hat{A}_i$ is the group-normalized advantage. The policy $\pi_{\mathrm{ref}}$ is the frozen SFT reference; $\mathcal{B}$ is a training batch. The start/end tokens $t_{\mathrm{start}}^{\mathrm r},t_{\mathrm{end}}^{\mathrm r}$ and $t_{\mathrm{start}}^{\mathrm s},t_{\mathrm{end}}^{\mathrm s}$ delimit reasoning and solution sections, and $\mathcal V_{\mathrm{phys}}$ is the predefined physical vocabulary. The total reward $R(y)$ is a weighted sum of three distinct physical alignment components: $R(y) = R_{\mathrm{fmt}}(y) + R_{\mathrm{sem}}(y, y^*) + R_{\mathrm{vocab}}(y)$.

\begin{algorithm}[tb]
   \caption{Group Tactile Policy Optimization (GTPO)}
   \label{alg:gtpo}
   \begin{algorithmic}[1]
   \STATE {\bfseries Input:} Dataset $\mathcal{D}$, Initial Policy $\pi_\theta$, Reference Model $\pi_{\mathrm{ref}}$, Semantic Encoder $E(\cdot)$, Group Size $G$, Learning Rate $\eta$, KL Coefficient $\beta$, Clip Ratio $\epsilon$.
   \STATE Initialize $\pi_{\mathrm{ref}} \leftarrow \pi_\theta$
   \REPEAT
   \STATE Sample a batch of visual prompts and ground truths $\mathcal{B} \sim \mathcal{D}$
   \STATE Update old policy $\pi_{\theta_{\text{old}}} \leftarrow \pi_\theta$
   \FOR{each visual prompt $x \in \mathcal{B}$}
       \STATE Sample $G$ outputs $\{y_1, y_2, \dots, y_G\} \sim \pi_{\theta_{\text{old}}}(\cdot | x)$
       \FOR{$i=1$ {\bfseries to} $G$}
           \STATE  \textbf{Rule-based} Formatting Reward: $r_{\text{fmt}} \leftarrow R_{\mathrm{fmt}}(y_i)$
           \STATE  \textbf{Rule-based} Vocabulary Reward: $r_{\text{vocab}} \leftarrow R_{\mathrm{vocab}}(y_i)$
           \STATE  \textbf{Model-based} Semantic Reward: $r_{\text{sem}} \leftarrow \text{clip}(1.5 \cdot (\cos(E(y_i), E(y^*)) + 0.1), 0, 2)$
           \STATE $r_i \leftarrow r_{\text{fmt}} + r_{\text{sem}} + r_{\text{vocab}}$
       \ENDFOR
       \STATE Compute Mean $\mu = \frac{1}{G}\sum_{j=1}^G r_j$ and Std $\sigma = \sqrt{\frac{1}{G}\sum_{j=1}^G (r_j - \mu)^2}$
       \FOR{$i=1$ {\bfseries to} $G$}
           \STATE $A_i \leftarrow (r_i - \mu) / (\sigma + \delta)$
       \ENDFOR
   \ENDFOR
   \STATE Optimize $\theta$ by maximizing the objective $\mathcal{J}_{\text{GTPO}}$:
   \FOR{each $y_i$ in group}
       \STATE \textbf{ratio} $\rho_i = \pi_\theta(y_i|x) / \pi_{\theta_{\text{old}}}(y_i|x)$, \textbf{surrogate} $L^{\text{clip}}_i = \min(\rho_i A_i, \text{clip}(\rho_i, 1-\epsilon, 1+\epsilon) A_i)$
   \ENDFOR
   \STATE Optimize $\theta$ by maximizing $\mathcal{J}$ in Eq.~\ref{eq:final_objective}
   \UNTIL{convergence}
   \STATE {\bfseries Output:} Optimized Physics-Aware Policy $\pi_\theta$
   \end{algorithmic}
\end{algorithm}

\paragraph{Reward Design.}
$R_{\mathrm{fmt}}$ enforces a structured Chain-of-Thought (CoT) process using reasoning and solution delimiters:
\begin{equation}
\begin{split}
    R_{\mathrm{fmt}}(y) = & \ 0.5 \cdot \mathbb{I}\left( t_{\text{start}}^{\text{r}} \in y \land t_{\text{end}}^{\text{r}} \in y \right) 
     + 0.5 \cdot \mathbb{I}\left( t_{\text{start}}^{\text{s}} \in y \land t_{\text{end}}^{\text{s}} \in y \right),
\end{split}
\label{eq:fmt_reward}
\end{equation}
$R_{\mathrm{sem}}$ aligns the generated physical description with ground truth physics in a semantic embedding space. We extract the text within the solution tags as $y_{\text{pred}}$ and compare it with the ground truth $y^*$. Using a pre-trained sentence encoder $E(\cdot)$, we compute the cosine similarity $s_{\cos}$. To penalize hallucinations and reward high-fidelity descriptions, we apply a clamped linear transformation:
\begin{equation}
    s_{\cos} = \frac{E(y_{\text{pred}}) \cdot E(y^*)}{\|E(y_{\text{pred}})\| \|E(y^*)\|}, \quad R_{\mathrm{sem}}(y) = \mathrm{clip}\left( 1.5 \cdot (s_{\cos} + 0.1), \: 0.0, \: 2.0 \right).
\end{equation}
Finally, $R_{\mathrm{vocab}}$ measures the density of valid physical terms in the prediction to prevent the generation of useless descriptions:
\begin{equation}
    R_{\mathrm{vocab}}(y) = 0.5 \cdot \frac{\sum_{w \in W(y_{\text{pred}})} \mathbb{I}(w \in \mathcal{V}_{\text{phys}})}{|W(y_{\text{pred}})|},
\end{equation}
where $W(y_{\text{pred}})$ denotes the set of words in the generated solution.

\subsection{Semantic-Driven Data Curation and Visual Prior}
\label{sec:alignment}
Real-world visual-tactile datasets often suffer from severe spatial misalignment due to mechanical tolerances. Spatial uncertainty can blur the outputs of deterministic pixel-wise regression and displace details in stochastic synthesis.

\paragraph{DINOv2 Semantic Filtering.}
To resolve this, we employ DINOv2~\citep{oquab2023dinov2}, a foundation model highly sensitive to material textures, to compute the semantic correlation between visual crops and tactile readings. Let $f_{\text{DINO}}(\cdot)$ denote the CLS token feature extraction of DINOv2. For each data pair $(I_v, I_t)$, we compute the semantic similarity score:
\begin{equation}
    S_{\text{DINO}} = \frac{f_{\text{DINO}}(I_v) \cdot f_{\text{DINO}}(I_t)}{\|f_{\text{DINO}}(I_v)\|_2 \|f_{\text{DINO}}(I_t)\|_2}.
\end{equation}
We use CLS features from \texttt{facebook/dinov2-base}, rank calibrated RGB--tactile pairs by this cosine score, and discard the bottom 30\%. This feature-based heuristic is applied before the split to all pairs; it does not use target labels or evaluation feedback, but it shapes the retained test distribution. Cross-modal DINO similarity is not a calibrated geometric alignment measure.

\paragraph{Depth Isolation and Calibration.}
Concurrently, we explicitly model the transformation $\mathcal{T}: \mathcal{I}_v \to \mathcal{I}_t$ to map the visual observation to the tactile sensor space using center-crop operations and rotations. We use the pretrained relative-depth model \texttt{Depth-Anything-V2-Large-hf}~\citep{yang2024depth}. Its output is min--max normalized per image to $[0,255]$, stored as uint8, and scaled to $[0,1]$ for ControlNet. This is a macro-geometric prior, not calibrated metric depth or the contact-induced elastomer deformation field.

\subsection{Physics-Driven Multi-Modal Tactile Image Generation}
\label{sec:synthesis}
To synthesize high-fidelity tactile responses that respect both macroscopic geometry and microscopic material properties, we adopt Stable Diffusion XL (SDXL)~\citep{podell2023sdxl} as our generative backbone. Unlike earlier models, SDXL features dual text encoders (CLIP ViT-L and OpenCLIP ViT-bigG), which significantly enhances its comprehension of complex GTPO-generated physical attributes. 

\paragraph{Four-Channel Geometric-Aware Input Representation.}
We augment the aligned RGB prior $I_{\mathrm{rgb}}$ with its corresponding depth map $I_{\mathrm{depth}}$ to form a 4-channel control tensor $\mathbf{C} \in \mathbb{R}^{H \times W \times 4}$. We modify the first convolutional layer of the SDXL ControlNet to accommodate this input. To preserve the pretrained visual feature extraction capabilities, we employ a channel-wise weight inheritance strategy. Let $\mathbf{W}_{\mathrm{old}} \in \mathbb{R}^{K \times 3 \times S \times S}$ be the original weights. The new weights $\mathbf{W}_{\mathrm{new}} \in \mathbb{R}^{K \times 4 \times S \times S}$ are initialized as:
\begin{equation}
    \mathbf{W}_{\mathrm{new}}^{(c)} = 
    \begin{cases} 
    \mathbf{W}_{\mathrm{old}}^{(c)} & \text{if } c \in \{0, 1, 2\}, \\
    \frac{1}{3} \sum_{j=0}^{2} \mathbf{W}_{\mathrm{old}}^{(j)} & \text{if } c = 3,
    \end{cases}
\label{eq:weight_init}
\end{equation}
This initialization ensures that the depth channel starts with a neutral but feature-aware contribution. The SDXL VAE, U-Net, and text encoders remain frozen; only ControlNet is trained.

\paragraph{Physically-Grounded Training Strategy.}
To prevent the model from overfitting to the non-contact background, we propose a \textbf{Contact-Aware Loss Reweighting} strategy. We generate a spatial weight mask $\mathbf{M}$ based on the depth prior $I_{\mathrm{depth}} \in [0, 1]$:
\begin{equation}
    \mathbf{M}_{u,v} = \begin{cases} 
    \lambda_{\mathrm{boost}} & \text{if } I_{\mathrm{depth}}(u,v) > \tau, \\
    1.0 & \text{otherwise},
    \end{cases}
\end{equation}
where $\tau=0.5$ and $\lambda_{\mathrm{boost}}=5.0$. The depth mask is resized to latent resolution. The diffusion training objective incorporates this mask to focus gradient descent on reconstructing fine-grained deformation details:
\begin{equation}
    \mathcal{L} = \mathbb{E}_{\mathbf{z}_t, t, \mathbf{c}_{\mathrm{txt}}, \mathbf{C}, \boldsymbol{\epsilon}} \left[ \left\| \mathbf{M} \odot \left(\boldsymbol{\epsilon} - \boldsymbol{\epsilon}_\theta(\mathbf{z}_t, t, \mathbf{c}_{\mathrm{txt}}, \mathbf{C})\right) \right\|_2^2 \right],
\end{equation}
where $\odot$ denotes element-wise multiplication, $\mathbf{z}_t$ is the noisy latent, and $\boldsymbol{\epsilon}_\theta$ is the predicted noise. This forces the model to focus gradient descent steps on reconstructing deformation details.

Furthermore, we implement a \textbf{Classifier-Free Guidance with Text Dropout}, randomly dropping the physics-aware text condition $\mathbf{c}_{\mathrm{txt}}$ with a probability $p=0.15$ during training. Finally, a \textbf{Physics-Prompt Normalization} protocol is applied to parse the LMM reasoning output into a structured set of concise physical tags.

% ==============================================================================
% 4. Experiments
% ==============================================================================
\section{Experiments}
\label{sec:experiments}

\subsection{Implementation Details}
\label{sec:exp_implementation}

We train PhysTacGen on SSVTP, which contains 4,587 original visual--tactile pairs with human material descriptions. DINOv2 curation retains 70\% of calibrated pairs before the train/test split. Paired crop/rotation augmentation yields approximately 13k training instances, and SSIM/LPIPS use approximately 1.2k held-out processed test instances; the separate reasoning benchmark has 452 pairs. All compared image-generation methods use the same retained split. The GTPO model is Qwen3-VL-8B-Instruct with group size 4, learning rate $2\times10^{-6}$, KL coefficient 0.05, clip ratio 0.2, and 1,000 training steps. The format, semantic, and vocabulary reward scales are 1.0, 1.5, and 0.5. Its vision encoder, vision--language adapter, and output head are frozen, while LoRA trains language-model attention and MLP modules; the reference policy is the frozen SFT model. For the SDXL generator, RGB and relative depth form the four-channel ControlNet condition. We train only ControlNet on 512$\times$512 images with AdamW at $10^{-5}$, fp16, batch size 1 with gradient accumulation 8, 10,000 steps, and 15\% text dropout. Final GTPO and generator runs used NVIDIA A100 GPUs; RTX 3090/4090 systems were used only for pilot work. SSIM uses aligned RGB images with \texttt{data\_range=255}; LPIPS uses VGG with inputs in $[-1,1]$. We report legacy nFD, computed by dividing the Fréchet feature distance by the evaluation-set size,which is not standard FID.

\begin{figure*}[ht]  
\vskip 0.2in  
\begin{center}    
\centerline{\includegraphics[width=\textwidth]{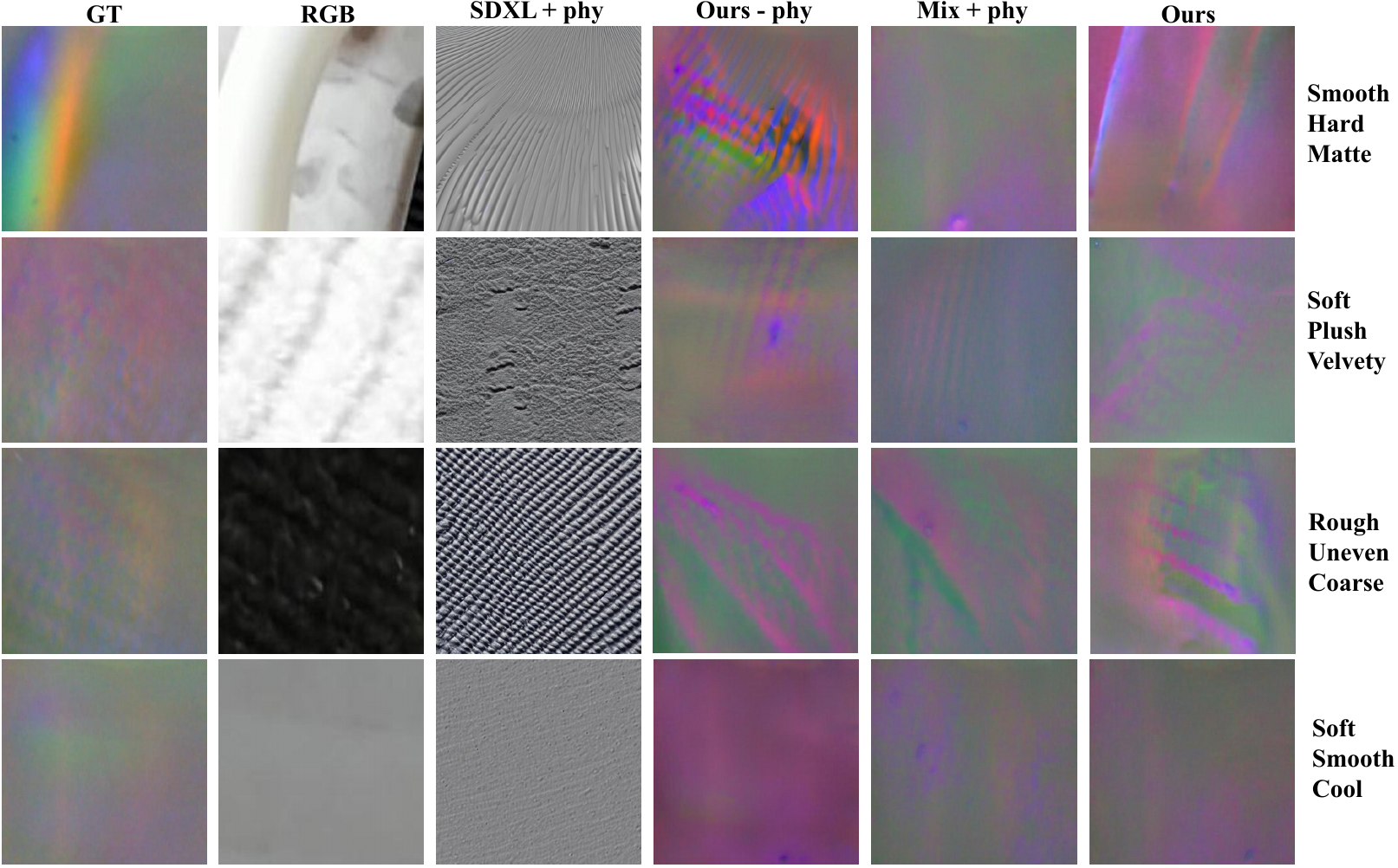}}    
\caption{\textbf{Qualitative Synthesis Results.} Visualisation of generated tactile images across different settings. SDXL (Baseline) fails to ground the contact location. The Mix-Fusion and four-channel variants change both fusion and loss and therefore do not isolate a single cause of their visual differences. The examples illustrate generated contact appearance under the original conditions.}   
\label{fig:visualization}  
\end{center}  
\vskip -0.2in
\end{figure*}

\subsection{Visual-to-Tactile Synthesis Quality}
\label{sec:exp_generation}

We compare PhysTacGen with three external baselines and the original generator variants on the same curated SSVTP split. Pix2Pix~\citep{isola2017image} uses U-Net-256/PatchGAN at 512$\times$512 for 200 epochs; Pix2Pix-Turbo~\citep{parmar2024one} uses one-step SD-Turbo at 512$\times$512 for 6,000 steps with a fixed generic prompt. VisGel~\citep{li2019connecting} uses the query RGB plus a reference RGB--tactile pair retrieved using RGB features from the training split only; query tactile data are never accessed. Because SSVTP lacks VisGel's original temporal/no-contact reference, this baseline is an adaptation.

\begin{table}[t]
  \caption{External baselines and original generator variants. All rows use the same retained SSVTP split. The last column is the legacy Fr\'echet statistic divided by evaluation-set size (legacy nFD), \emph{not} FID.}
  \label{tab:ablation_results}
  \centering\small\setlength{\tabcolsep}{4pt}
  \begin{tabular}{lccc}
  \toprule
  Method and condition & SSIM $\uparrow$ & LPIPS $\downarrow$ & legacy nFD $\downarrow$ \\
  \midrule
  Pix2Pix / RGB & 0.106 & 0.819 & 0.829 \\
  Pix2Pix-Turbo / RGB + generic prompt & 0.417 & 0.692 & 0.628 \\
  VisGel (adapted) / RGB + train reference & 0.715 & 0.629 & 0.433 \\
  SDXL baseline / physical text only & 0.190 & 0.859 & 0.838 \\
  Mix-Fusion / RGB + depth, physical text & 0.938 & 0.617 & \textbf{0.241} \\
  Four-channel / RGB + depth, generic text & 0.908 & 0.637 & 0.349 \\
  \textbf{PhysTacGen} / RGB + depth, GTPO text & \textbf{0.964} & \textbf{0.559} & 0.278 \\
  \bottomrule
  \end{tabular}
\end{table}

\begin{figure}[h]
  \centering
  \includegraphics[width=\linewidth]{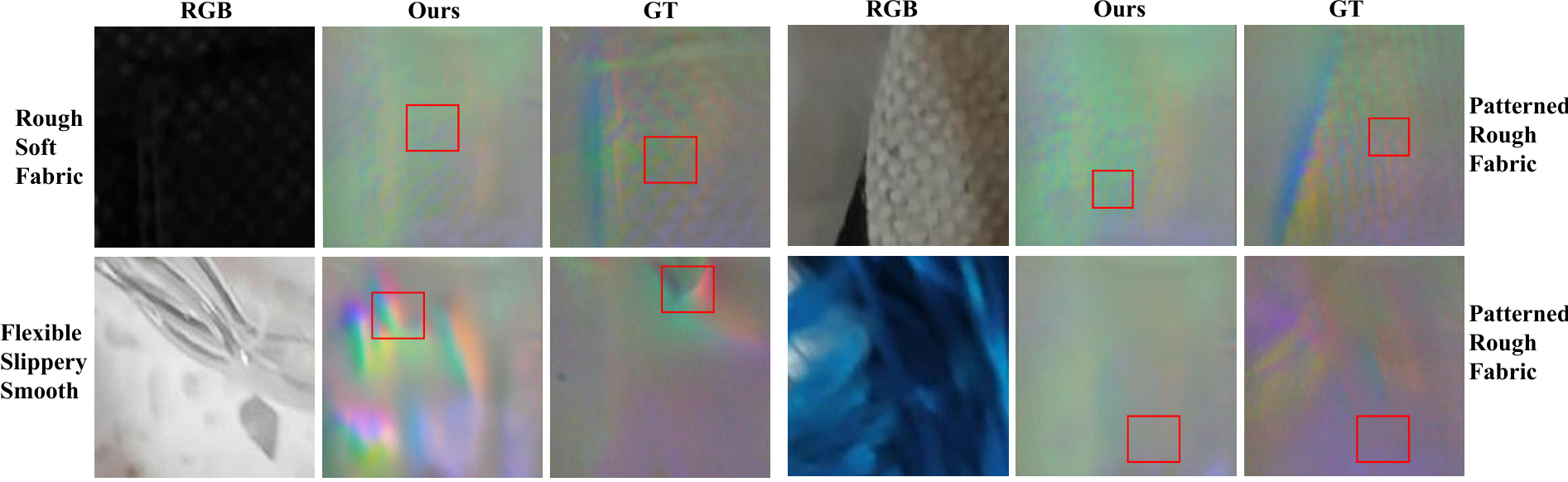} % 在此处放入排版好的 RGB, GT, Result 对比图
  \caption{\textbf{Material and Textural Consistency under Spatial Misalignment.} Red boxes highlight areas for qualitative texture comparison.The details in the areas outlined in red in each set of images are similar.} 
  \label{fig:material_consistency}
\end{figure}

Table~\ref{tab:ablation_results} shows the highest SSIM and lowest LPIPS for PhysTacGen among the original variants and the evaluated external baselines; Mix-Fusion obtains the lowest legacy nFD. Since Mix-Fusion changes both fusion and loss, that row cannot isolate either effect. The generic-text variant tests the impact of semantic text with the same four-channel architecture and contact-aware training.

\paragraph{Matched prompt-source comparison.}
For a matched comparison of prompt sources, we train three separate ControlNet generators from the same pretrained initialization.Classifier, SFT, and GTPO prompts are each used during both training and test time for their own model. The DINO-retained pairs, split, RGB-depth inputs, architecture, contact-aware loss, optimizer, training steps, random seed, and inference settings are held fixed. The seven-label classifier is trained from original SSVTP human descriptions, not distilled from GTPO; its vocabulary is textured, hard, smooth, rough, soft, flexible, and fabric. Since captions may omit valid properties, absent labels need not be true negatives.

\begin{table}[t]
  \caption{Final matched end-to-end prompt-source comparison. Each prompt source trains and tests a separate generator under the same conditions; legacy nFD is not standard FID.}
  \label{tab:matched_prompts}
  \centering\small
  \begin{tabular}{lccc}
  \toprule
  Prompt source & SSIM $\uparrow$ & LPIPS $\downarrow$ & legacy nFD $\downarrow$ \\
  \midrule
  7-label classifier & 0.919 & \textbf{0.557} & 0.305 \\
  SFT & 0.948 & 0.570 & \textbf{0.255} \\
  GTPO & \textbf{0.964} & 0.559 & 0.278 \\
  \bottomrule
  \end{tabular}
\end{table}

Relative to SFT, GTPO improves SSIM from 0.948 to 0.964 and LPIPS from 0.570 to 0.559; SFT has lower legacy nFD. Relative to the classifier, GTPO improves SSIM and legacy nFD, while the classifier has LPIPS lower by 0.002. A simplified lexical consistency check counts descriptions containing both hard and soft or both smooth and rough. Contradiction rates are 7.6\% for the classifier, 3.2\% for SFT, and 1.3\% for GTPO.  Figure~\ref{fig:material_consistency} offers qualitative examples of residual spatial offsets.

\subsection{Downstream Application: Synthetic-Tactile Substitution for Proxy Force-Coefficient Prediction}
\label{sec:exp_downstream}

The force-coefficient target $y\in[0.1,0.9]$ is assigned heuristically from normalized SSVTP material descriptions, rather than measured with a force sensor. In the fixed attribute mapping, ``slippery'' adds 0.25 and ``soft'' subtracts 0.15. A late-fusion ResNet-18 $\pi(I_v,I_t)$ is trained on real RGB--real tactile pairs with a 90:10 split. At test time we compare RGB-only, RGB plus generated tactile, and RGB plus real tactile (oracle). This tests substitution of a synthetic observation on the same proxy task; it does not test simulator-to-robot transfer or closed-loop manipulation.

\begin{table}[t]
  \caption{Test MSE for the heuristic force-coefficient prediction proxy. Sample SD describes variation across from two random seeds; the last column reports their mean and sample standard deviation.}
  \label{tab:downstream_results}
  \centering\small
  \begin{tabular}{lccc}
  \toprule
  Input modality & Run 1 & Run 2 & Mean $\pm$ sample SD \\
  \midrule
  RGB only & 0.144 & 0.147 & $0.1455\pm0.0021$ \\
  RGB + generated tactile & 0.132 & 0.134 & $0.1330\pm0.0014$ \\
  RGB + real tactile (oracle) & 0.130 & 0.132 & $0.1310\pm0.0014$ \\
  \bottomrule
  \end{tabular}
\end{table}

Generated-tactile input improves on RGB-only in both runs, and its MSE is 0.002 above the real-tactile oracle in each. As the target is built from material attributes also used in the semantic pipeline, these results indicate consistency on a proxy task, not measured-force utility. Two final generator models trained on independently selected 9:1 folds differed by less than 5\% in SSIM/LPIPS.

\subsection{Comprehensive Evaluation of Physical Reasoning (GTPO)}
\label{sec:exp_gtpo_eval}

 We assess GTPO-generated physical descriptions against SFT and a seven-label classifier on the standardized 452-pair benchmark, using lexical metrics, an auxiliary automatic judge, and blinded human preference.

\paragraph{Data Curation and Vocabulary Standardization.}
The raw SSVTP dataset contains unstructured human descriptions often polluted by purely visual adjectives or vague terms. To construct a high-quality ground truth for physical reasoning, we implemented a rigorous text normalization pipeline: (1) \textbf{Visual Noise Filtering:} We constructed a blacklist to discard non-tactile descriptors. (2) \textbf{Physical Attribute Normalization:} We mapped synonymous adjectives to a standardized physics vocabulary. For instance, terms like ``rigid", ``stiff", and ``ceramic" are mapped to the canonical label \textit{``hard"}. This process yielded a refined dataset of 452 unique object-material pairs for testing.

\paragraph{Evaluation Metrics.}
We employ BLEU~\citep{papineni2002bleu} and ROUGE~\citep{lin2004rouge} to measure linguistic overlap with expert annotations, while Perplexity (PPL) is used to measure fluency. As an auxiliary automatic measure, we use a \textbf{VLM Physics Judge}. We prompt GPT-4o to act as an expert which reviews the generated descriptions for apparent physical plausibility and outputs a \textit{Physics Plausibility Score} between 1-5. Furthermore, we extract the physical keywords from the tags and compute the Semantic Embedding Similarity (SES) using \texttt{all-MiniLM-L6-v2} to capture objective vector-space alignment.

\begin{table*}[t]
  \caption{Description evaluation on 452 pairs. The human column is a blinded forced choice among classifier, SFT, and GTPO descriptions by 22 participants, with randomized presentation and one choice per image.}
  \label{tab:reasoning_results}
  \centering\scriptsize\setlength{\tabcolsep}{3pt}
  \begin{tabular}{lccccccc}
  \toprule
  Method & SES $\uparrow$ & BLEU $\uparrow$ & ROUGE $\uparrow$ & PPL $\downarrow$ & Judge $\uparrow$ & Human $\uparrow$ \\
  \midrule
  7-label classifier & \textbf{0.732} & \textbf{29.3} & \textbf{52.7}& 12.9 & 2.94 & 16\% \\
  SFT Qwen & 0.461 & 18.4 & 32.7& 14.3 & 3.42 & 27\% \\
  GTPO Qwen & 0.586 & 26.8 & 44.5  & \textbf{10.8} & \textbf{4.65} & \textbf{57\%} \\
  \bottomrule
  \end{tabular}
\end{table*}

\paragraph{Results Analysis.} 
As Table~\ref{tab:reasoning_results} shows, the closed-vocabulary classifier leads SES, BLEU, and ROUGE; this favors frequent canonical words. GTPO has the lowest reported PPL, highest auxiliary judge score (4.65), and highest human preference (57\%). The study used 22 participants; every evaluated image was presented to all 22, with descriptions in randomized order and no ties. These measures support the quality of GTPO descriptions but do not directly validate contact mechanics. The matched generator table additionally shows a modest SSIM and LPIPS advantage over SFT, with trade-offs against the classifier.

% ==============================================================================
% Limitations and Broader Impacts
% ==============================================================================
\section{Limitations and Broader Impacts}
\label{sec:limitations}

\paragraph{Limitations and Future Work.}

The generation results are limited to one curated SSVTP dataset. DINO filtering occurred before the split, shaping the retained test distribution, and direct DINO feature similarity across camera and optical tactile images is a heuristic. Homogeneous surfaces and residual spatial offsets can still cause errors. The two-seed and two-fold checks show consistency but not formal significance or cross-sensor generalization. The force-coefficient labels are heuristic proxies, so the downstream result cannot demonstrate measured force or real robot transfer. The distributional comparison uses legacy nFD rather than standard FID/KID. The heuristic coefficient mapping is specified only in part, and the current experiments do not isolate every component or assess cross-sensor generalization.

\paragraph{Broader Impacts.}

This study investigates synthetic optical tactile observations as a potential supplement to costly paired data collection. Generated outputs can be inaccurate, especially outside the curated distribution; applications involving force control or safety-critical contact require validation with calibrated sensors and real robots.

\section{Conclusion}
\label{sec:conclusion}

We presented \textbf{PhysTacGen}, which combines GTPO-derived material descriptions, DINOv2-based pair curation, relative-depth conditioning, and an SDXL ControlNet for visual-to-optical-tactile synthesis. On the curated SSVTP test split, PhysTacGen achieves an SSIM of 0.964 and an LPIPS of 0.559. In a matched comparison of separately trained generators, GTPO prompts yield the highest SSIM, while seven-label-classifier prompts achieve a marginally lower LPIPS. In a blinded study with 22 participants, GTPO descriptions receive the highest preference rate of 57\%. Generated tactile inputs also reduce test MSE relative to RGB-only inputs in both runs of the attribute-derived force-coefficient proxy task. Together, these results demonstrate PhysTacGen's optical tactile image synthesis performance and its utility for the evaluated proxy task on curated SSVTP data.

% ==============================================================================
% References and Checklist
% ==============================================================================
% \section*{References}
% \medskip
{
\small
\bibliographystyle{plainnat}
% Add these verified reference keys to the existing ref.bib before compiling:
% isola2017pix2pix, parmar2024pix2pixturbo, li2019visgel,
% gao2023controllable, shao2024deepseekmath, yang2024depthanythingv2.
\bibliography{ref} % 确保你有对应的 ref.bib 文件
}

% \bibliographystyle{plainnat}
% \bibliography{ref} % <-- 请在这里修改为你的.bib文件名 (去掉.bib后缀)

% The checklist is omitted from this PDF as requested.
% \newpage
% \input{checklist.tex}

\end{document}